\documentclass[letterpaper]{article} 
\usepackage[preprint]{aaai2027}  
\usepackage[hyphens]{url}  
\usepackage{graphicx} 
\usepackage{natbib}  
\usepackage{caption} 
\usepackage{algorithm}
\usepackage{algorithmic}
\usepackage{amsmath,amssymb}
\usepackage{xcolor}

\usepackage{newfloat}
\usepackage{listings}
\DeclareCaptionStyle{ruled}{labelfont=normalfont,labelsep=colon,strut=off} 
\floatstyle{ruled}
\newfloat{listing}{tb}{lst}{}
\floatname{listing}{Listing}

\usepackage{booktabs}
\usepackage{multirow}

\title{Residual Flow Matching with Dynamic Cross-Interaction for 3D Multi-Person Motion Prediction}
\author{
    Wei Wei\textsuperscript{\rm 1}\equalcontrib,
    Yinyuan Zhao\textsuperscript{\rm 1}\equalcontrib,
    Ruixuan Yu\textsuperscript{\rm 1}\corresponding
}
\affiliations{
    \textsuperscript{\rm 1}Shandong University\\
    202300620418@mail.sdu.edu.cn, 
    202300810128@mail.sdu.edu.cn, 
    yuruixuan@sdu.edu.cn
}

\begin{document}

\maketitle

\begin{abstract}
3D multi-person motion prediction requires modeling both individual kinematics and inter-person interactions. While Flow Matching is effective for multi-hypothesis generation to improve prediction accuracy, directly predicting skeletal sequences from pure noise often compromises structural consistency and introduces unreliable cross-agent interactions during early noise-dominated integration steps. To address this, we propose a Prior-Guided Residual Flow Matching framework. First, a Deterministic Coarse Prior (DCP) establishes a kinematic anchor, formulating the generative process as a conditional flow over motion residuals to simplify the generative objective and preserve structural stability. Second, a Dynamic Cross-Interaction (DCI) mechanism temporally synchronizes inter-agent message-passing with the integration progress, ensuring the extraction of reliable social contexts and improving multi-person motion fidelity. Finally, a decoupled joint-motion architecture with bidirectional fusion effectively preserves fine-grained kinematic coherence. Extensive experiments demonstrate that our approach achieves state-of-the-art prediction accuracy across multiple datasets. Code is available at https://github.com/Wei-Wei-a/Residual-Flow-Matching-with-Dynamic-Cross-Interaction-for-3D-Multi-Person-Motion-Prediction.

\end{abstract}

\section{Introduction}

3D multi-person motion prediction aims to forecast future skeletal configurations from historical observations, providing essential representations for downstream applications such as autonomous navigation and human-robot interaction. Effectively modeling these dynamics presents intertwined spatial and dynamic challenges. Spatially, generating plausible multi-person sequences requires integrating individual kinematics with inter-agent spatial dependencies. Dynamically, human interactions are inherently stochastic, and identical histories often lead to diverse yet  plausible future trajectories~\cite{yuan2020dlow,xu2023dummf,sun2024eccv-comusion}.

While conventional deterministic methods leverage recurrent, graph-based, Transformer, or MLP architectures to explicitly model spatio-temporal dependencies~\cite{fragkiadaki2015recurrent,mao2019learning,xu2023jointr,zheng2025empmp}, their optimization for a single trajectory forces convergence toward the conditional mean of the data distribution, restricting capacity to capturing multimodal future interactions. Then, continuous-time generative models, such as Diffusion and Flow Matching, are  adopted~\cite{chen2023humanmac,sun2024eccv-comusion}. Nevertheless, synthesizing high-dimensional 3D sequences directly from an unstructured noise prior imposes an  optimization burden. Learning the underlying skeletal structure and diverse future dynamics simultaneously from scratch often leads to kinematic distortions and history-inconsistent motions~\cite{barquero2023belfusion,yuan2023physdiff}. To alleviate this, we decouple the forecasting objective by first establishing a deterministic kinematic prior, thereby constraining the generative model to synthesize the motion residuals.


Beyond individual kinematics, accurate multi-person forecasting necessitates modeling inter-agent interactions~\cite{xu2023jointr,xiao2024iaformer}. When adapting spatial fusion to continuous-time generative models, existing approaches typically apply cross-person aggregation uniformly throughout denoising, without modulating interaction strength based on integration progress~\cite{tanke2023socialdiffusion}. However, early-stage states in Ordinary Differential Equation (ODE) integration are highly noise-corrupted. Applying unconstrained spatial message-passing here compels the network to extract social contexts from unreliable representations, introducing spurious correlations and degrading motion fidelity. Although timestep embeddings are standard for temporal conditioning~\cite{tevet2023mdm,yuan2023physdiff}, explicitly aligning interaction magnitude with integration progress remains unexplored. To address this, we introduce a time-varying mechanism that progressively amplifies inter-agent message-passing, mitigating early-stage noise interference.

In this work, we propose \textit{\textbf{Prior-Guided Residual Flow Matching}}, a framework for 3D multi-person motion prediction that decomposes the forecasting objective into a deterministic prior and generative residuals. Our main contributions are: 
 \textit{\textbf{First}}, we introduce a two-stage generative paradigm that integrates a deterministic prior with a continuous-time residual flow. This joint formulation effectively mitigates the structural distortions inherent in pure-noise synthesis. 
 \textit{\textbf{Second}}, we design a hierarchical dual-branch representation to model motion at joint, individual, and social levels. It isolates local skeletal topology from inter-agent interactions, preserving structural integrity during generation. 
 \textit{\textbf{Third}}, we propose the Dynamic Cross-Interaction (DCI)  that synchronizes spatial message-passing within ODE integration progress. By temporally gating social attention, it suppresses early-stage noise interference, facilitating  extraction of reliable interactive contexts. 
 \textit{\textbf{Finally}}, extensive experiments demonstrate that our method achieves state-of-the-art performance across diverse multi-person motion benchmarks.


\section{Related Work}

\paragraph{Multi-Person Motion Forecasting.}

Human motion forecasting has evolved from deterministic single-person modeling~\cite{li2018convoseq,mao2020hisrep,aksan20213dv} to complex multi-person scenarios. Recent frameworks employ various spatio-temporal architectures to capture inter-person social contexts and expressive whole-body movements~\cite{adeli2021tripod,wang2021mrt,peng2023tbiformer,xu2023jointr,ding2024aaai-expressive,zheng2025empmp}. While these  methods  model historical dependencies, they commonly yield a single future trajectory. To capture the multimodal nature of human behavior, subsequent studies introduced stochastic motion prediction. Latent-variable models, such as VAEs and GANs, are  utilized to generate diverse futures~\cite{barsoum2018hpgan,yuan2020dlow,aliakbarian2021contextually,mao2021gsps}, and have also been adapted for semantic-conditioned motion synthesis~\cite{petrovich2021actor,petrovich2022temos,guo2022textmotion,zhang2024motiondiffuse}. However, modeling the high-dimensional distributions for 3D articulated skeletons over extended horizons remains challenging.

Recently, continuous-time generative models, including Diffusion~\cite{tevet2023mdm,chen2023humanmac} and Flow Matching~\cite{lipman2023flow}, have been applied to human trajectory and pose forecasting~\cite{saadatnejad2023deposit,fu2025moflow,tian2025prediflow}. To improve generation quality, recent efforts also incorporate physical constraints and structural regularizations~\cite{barquero2023belfusion,yuan2023physdiff,sun2024eccv-comusion}. Nevertheless, in standard formulations, mapping an isotropic noise  directly to high-dimensional 3D skeletal sequences requires the model to concurrently learn local skeletal topology and stochastic temporal variations, often complicating the modeling process. In this work, we restructure the generative paradigm by formulating the continuous-time flow over motion residuals. Anchored by a deterministic coarse prior, we explicitly decouple structural constraints from stochastic multi-agent dynamics, enabling the generative model to focus entirely on capturing the complex, multimodal distribution of future interactions.

\paragraph{Dynamic Social Interaction in Generative Models.}

Modeling inter-person dependencies is a fundamental challenge in 3D multi-person motion forecasting, where the network must concurrently capture individual articulated joint dynamics and complex social interactions. Recent generative frameworks, such as Social Diffusion~\cite{tanke2023socialdiffusion} and latent-variable multi-person models~\cite{xu2023dummf}, facilitate joint generation by embedding cross-person aggregation directly into the network architecture. While state-of-the-art diffusion models effectively utilize timestep embeddings to condition the denoising process for individual motions (e.g., MDM~\cite{tevet2023mdm}, TransPhase~\cite{au2025transphase}), existing multi-agent architectures typically employ a time-invariant interaction topology. This static architectural design largely inherits interaction paradigms from deterministic modeling, lacking explicit temporal modulation of spatial message-passing based on the integration progress.

Under the continuous-time generative formulation, applying time-invariant spatial interaction can lead to suboptimal integration trajectories. During the early stages of the generative process (e.g., solving the Ordinary Differential Equation), the latent states are predominantly isotropic noise. Enforcing dense cross-person attention on these uninformative representations compels agents to exchange noise rather than meaningful kinematic features. For high-dimensional 3D skeletons, this early-stage interference propagates spurious correlations across individuals, often resulting in joint distortions and structural homogenization. While timestep-aware scheduling has been successfully explored to prevent noise interference when enforcing physical constraints (e.g., PhysDiff~\cite{yuan2023physdiff}), explicitly gating the inter-person interaction mechanism in 3D multi-agent contexts remains underexplored. To resolve this, we introduce the Dynamic Cross-Interaction (DCI) mechanism. By restricting spatial message-passing during the noise-dominated phase and progressively restoring social dependencies as the latent states stabilize, DCI effectively mitigates early-stage noise interference while capturing reliable interactive contexts.

\section{Method}
\subsection{Problem Formulation and Framework Overview}
\label{sec:overview}

\begin{figure*}
    \centering
    \includegraphics[width=1\linewidth]{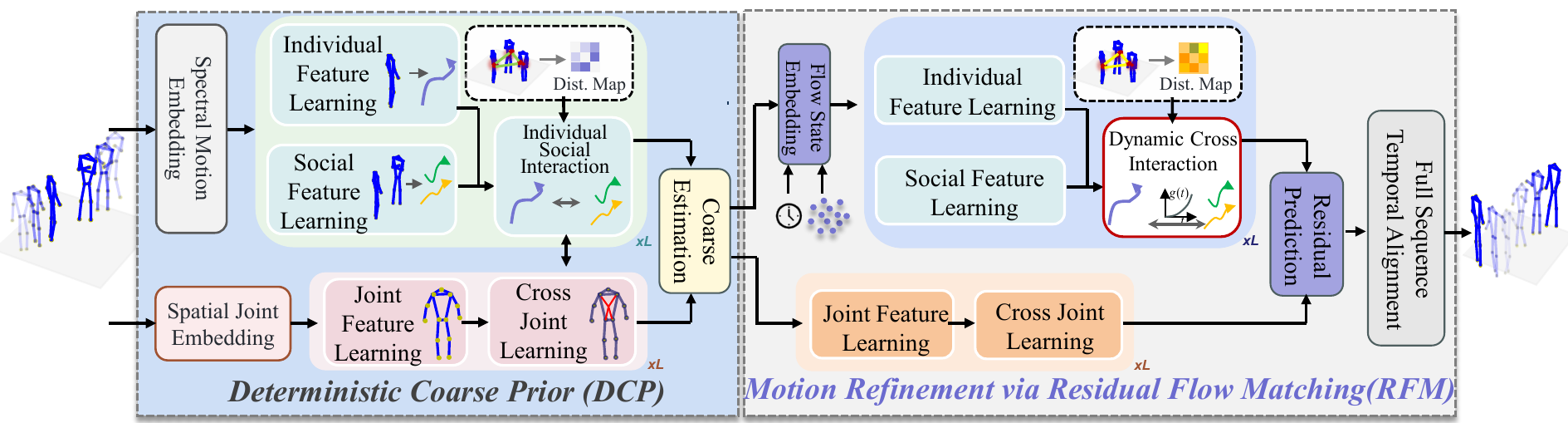}
    \caption{Overview of the proposed framework. The Deterministic Coarse Prior (DCP) module extracts a  motion prior by integrating hierarchical representations. The Residual Flow Matching (RFM) module then refines this prior into future predictions, utilizing Dynamic Cross-Interaction (DCI) to adaptively modulate global-to-local feature interactions.}
    \label{fig:pipeline}
\end{figure*}

Let $\mathbf{X} \in \mathbb{R}^{3J \times P \times T_h}$ denote the observed 3D  history motion of $P$ interacting individuals with $J$ joints over $T_h$ frames. We aim to predict  $K$ plausible future sequences $\{\mathbf{Y}_k^*\}_{k=1}^K \in \mathbb{R}^{3J \times P \times T_f}$ over $T_f$ frames. 
While continuous-time generative models (e.g., Flow Matching) offer a powerful paradigm for capturing the inherent uncertainty and non-linear dynamics of human motion, directly synthesizing multi-person trajectories from isotropic noise presents challenges. Specifically, mapping pure noise to high-dimensional skeletal sequences often struggles to maintain spatial consistency, leading to structural distortions in the generated poses. Furthermore, multi-person motion is intrinsically hierarchical; jointly encoding local skeletal topology and global social interactions without explicit separation often compromises individual motion fidelity. Moreover, in continuous-time generation, intermediate states at early ODE integration stages are heavily corrupted by noise. Extracting inter-agent social contexts from these unreliable states inevitably introduces spurious correlations.

To address these limitations, we propose \textit{Prior-Guided Residual Flow Matching}, a framework that decomposes the forecasting objective into a deterministic prior and generative residuals via three components (Figure~\ref{fig:pipeline}). \textbf{First}, a Deterministic Coarse Prior (DCP) module predicts an initial trajectory $\mathbf{Y}_c$ to condition the residual flow, effectively anchoring the generative process and enhancing prediction accuracy. \textbf{Second}, a hierarchical network models motion across joint, individual, and social levels. This decoupling isolates skeletal topology from global multi-agent interactions, mitigating the degradation of individual motion fidelity. \textbf{Finally}, a \textit{Dynamic Cross-Interaction (DCI)} mechanism temporally modulates inter-agent message passing. By gating social attention during early noise-dominated ODE phases, DCI facilitates the extraction of reliable interactive contexts.


\subsection{Deterministic Coarse Prior (DCP)}
\label{sec:dcp}

This module extracts a deterministic coarse prior from historical observations as reliable kinematic foundation. We first adopt a decoupled joint-motion design to  capture localized joint features and global multi-agent dynamics, which are then fused to predict the coarse future representation.

\paragraph{Spatial Joint Encoding.}
The spatial branch  explicitly model frame-wise skeletal topology. Given initial embeddings $\mathbf{Z}_J^{(0)} = \phi(\mathbf{X})$ extracted via a linear mapping $\phi$ on the 3D joint-wise coordinate, the features are sequentially refined across $L$ layers. Each layer conduct joint feature learning and cross joint learning,  capturing both independent joint dynamics and cross-joint structural dependencies:
\begin{equation}
\begin{split}
     \tilde{\mathbf{Z}}_J^{(l)} &= \mathbf{Z}_J^{(l-1)} + \varphi_{\mathrm{ind}}^{(l)}(\mathbf{Z}_J^{(l-1)}),\\
     \mathbf{Z}_J^{(l)} = &\tilde{\mathbf{Z}}_J^{(l)} + \mathcal{F}^{-1}\Big( \varphi_{\mathrm{mix}}^{(l)}\big(\mathcal{F}(\tilde{\mathbf{Z}}_J^{(l)})\big) \Big),
\end{split}
\end{equation}  
where $\varphi_{\mathrm{ind}}^{(l)}$ and $\varphi_{\mathrm{mix}}^{(l)}$ are MLPs. The operator $\mathcal{F}(\cdot)$ flattens the per-person joint features into a unified vector to model full-body correlations, which is reshaped back via $\mathcal{F}^{-1}(\cdot)$.

\paragraph{Temporal Motion Encoding.}
To effectively model long-horizon temporal dynamics and multi-agent interactions, the motion branch progressively learns the motion features through a stack of $L$ encoding layers. Specifically, following EMPMP~\cite{zheng2025empmp}, the historical sequence $\mathbf{X}$ is first transformed via a Discrete Cosine Transform (DCT) to initialize the motion embedding $\mathbf{Z}_M^{(0)}$. Within each of the $L$ layers, the global feature $\mathbf{Z}_M^{(l-1)}$ is explicitly branched into local individual features $\mathbf{Z}_{\mathrm{loc}}^{(l)}$ and global social states $\mathbf{Z}_{\mathrm{glo}}^{(l)}$ via linear mappings $\theta_{\mathrm{loc}}^{(l)}$ and $\theta_{\mathrm{glo}}^{(l)}$:
\begin{equation}
\begin{split}
    \mathbf{Z}_{\mathrm{loc}}^{(l)} &= \mathbf{Z}_M^{(l-1)} + \theta_{\mathrm{loc}}^{(l)}(\mathbf{Z}_M^{(l-1)}), \\ \mathbf{Z}_{\mathrm{glo}}^{(l)} = &\mathcal{H}(\mathbf{Z}_M^{(l-1)}) + \theta_{\mathrm{glo}}^{(l)}\big(\mathcal{H}(\mathbf{Z}_M^{(l-1)})\big),
\end{split}
\end{equation} 
where $\mathcal{H}(\cdot)$ reshapes the feature to process person and temporal dimensions jointly. To model multi-agent interactions, a bidirectional individual-social interaction mechanism is applied. The global context, augmented by an history inter-person distance embedding $\tau(\mathbf{D})$, modulates local features via learned affine parameters. Conversely, aggregated local features predict a translation to refine the global state:
\begin{equation}
\begin{split}
     \tilde{\mathbf{Z}}_{\mathrm{loc}}^{(l)} = \mathbf{Z}_{\mathrm{loc}}^{(l)} &+ \nu\Big( \mathbf{Z}_{\mathrm{loc}}^{(l)} \odot (1 + \mathbf{S}^{(l)}) + \mathbf{H}^{(l)} + \tau(\mathbf{D}) \Big), \\ \tilde{\mathbf{Z}}_{\mathrm{glo}}^{(l)} &= \mathbf{Z}_{\mathrm{glo}}^{(l)} + \nu\Big( \mathbf{Z}_{\mathrm{glo}}^{(l)} + \mathbf{G}^{(l)} \Big), 
\end{split}\label{eq:aff}
\end{equation}
where $\nu$ is layer normalization, $\odot$ is element-wise multiplication, and the affine parameters are generated by  MLPs: 
\begin{equation}
    \mathbf{S}^{(l)}, \mathbf{H}^{(l)} = \eta_{\mathrm{glo}}(\mathbf{Z}_{\mathrm{glo}}^{(l)}),~~\mathbf{G}^{(l)} = \eta_{\mathrm{loc}}(\mathcal{H}(\mathbf{Z}_{\mathrm{loc}}^{(l)})).
\end{equation}
Finally, the two branches are fused back into a unified representation for the next layer via $\mathbf{Z}_M^{(l)} = \tilde{\mathbf{Z}}_{\mathrm{loc}}^{(l)} + \alpha \mathcal{H}^{-1}\big(\tilde{\mathbf{Z}}_{\mathrm{glo}}^{(l)}\big)$, with $\alpha$ controlling the global contribution.

\paragraph{Cross-Branch Interaction and Coarse Estimation.} 
To model the dependencies between spatial joint kinematics and temporal motion patterns, we periodically (e.g., every four layers) introduce a bidirectional gated mechanism to exchange representations within the hierarchical encoding:
\begin{equation}
\begin{split}
        \mathbf{Z}_M^{\prime (l)} &= \mathbf{Z}_M^{(l)} + \sigma(\mathbf{Z}_M^{(l)} \mathbf{W}_{JM}) \odot \mathbf{Z}_J^{(l)}, \\ 
        \mathbf{Z}_J^{\prime (l)} &= \mathbf{Z}_J^{(l)} + \sigma(\mathbf{Z}_J^{(l)} \mathbf{W}_{MJ}) \odot \mathbf{Z}_M^{(l)},
\end{split}
\end{equation} 
where $\sigma$ is the Sigmoid activation and $\mathbf{W}$ denotes learnable projection matrices. After the $L$-layer encoding, the features from both branches are concatenated and projected into a unified latent representation $\mathbf{Z}_c$. An MLP head $\xi$ first maps $\mathbf{Z}_c$ to  residual DCT coefficients. These coefficients are then transformed back to the spatial domain via inverse DCT (iDCT) to yield the coarse spatial motion residual $\mathbf{R}_c$:
\begin{equation}
\begin{split}
    \mathbf{Z}_c = [\mathbf{Z}_M^{(L)}, &\mathbf{Z}_J^{(L)}] \mathbf{W}_c, \quad \mathbf{R}_c = \mathrm{iDCT}(\xi(\mathbf{Z}_c)),
\end{split}
\end{equation}
where $[,]$ is feature concatenation. The spatial residual $\mathbf{R}_c$ is anchored to the last observed pose $\mathbf{X}_{T_h}$ to formulate the deterministic coarse prediction $ \mathbf{Y}_c = \mathbf{R}_c + \mathbf{X}_{T_h} $.
Furthermore, to explicitly encode the relative position of the predicted motions, we compute a frame-wise $L_2$ distance matrix $\mathbf{D}_c$ of the spatial residuals $\mathbf{R}_c$.  The tuple $(\mathbf{Z}_c, \mathbf{R}_c, \mathbf{D}_c)$ serves as the deterministic prior for the subsequent refinement stage.

\subsection{Prior-Guided Residual Flow Refinement}
While the coarse stage provides a reliable kinematic prior, deterministic formulations exhibit limitations in modeling the complex non-linear dynamics of multi-agent interactions. To capture these intricate temporal dependencies, we introduce a conditional residual flow matching framework. Rather than learning an unconstrained vector field, we formulate the continuous-time probability flow over motion residuals. Specifically, given the ground-truth future motion $\mathbf{Y}$ and the last observed pose $\mathbf{X}_{T_h}$, the generative target is defined as the clean endpoint residual $\mathbf{R}^{(1)} = \mathbf{Y} - \mathbf{X}_{T_h}$.  During training, for a sampled time step $t \sim \mathcal{U}(0,1)$ and initial noise $\mathbf{R}^{(0)} \sim \mathcal{N}(\mathbf{0}, \sigma^2 \mathbf{I})$, the intermediate state is constructed via linear interpolation:
\begin{equation}
    \mathbf{R}^{(t)} = t \mathbf{R}^{(1)} + (1-t) \beta\mathbf{R}^{(0)},
\end{equation}  
where $\beta$ is a scaling parameter. This residual formulation simplifies the generative objective to modeling the residual distribution, facilitating future motion generation anchored by the coarse prior.
Furthermore, following recent flow-based practices~\cite{fu2025moflow}, we adopt an endpoint parameterization to directly predict the clean target $\mathbf{R}^{(1)}$.

\paragraph{Noise-Aware Motion Refinement via Dynamic Cross-Interaction.}
To effectively guide the generative process, the refinement network explicitly integrates the deterministic prior with the time-varying noisy states. For a given integration step $t$, the spatial noisy state $\mathbf{R}^{(t)}$ and the time variable $t$ are embedded and fused with the deterministic latent prior $\mathbf{Z}_c$ from the coarse stage. The initial motion feature is constructed as:
\begin{equation}
    \mathbf{H}_M^{(0)} = \mathrm{DCT}\Big( \psi_{\mathrm{in}}\big( \big[ \mathbf{Z}_c, \, \phi_r(\mathbf{R}^{(t)} + \phi_t(t) \big] \big) \Big),
\end{equation}
where $\phi_r$ and $\phi_t$ are linear projections, and $\psi_{\mathrm{in}}$ denotes a composite block of MLPs coupled with a frame-wise distance-guided cross-person multi-head attention module to explicitly capture spatial social interactions.

The initialized feature $\mathbf{H}_M^{(0)}$ is subsequently processed by $L$ refinement layers, sharing the local-global decoupling architecture of the coarse stage. As intermediate states near $t \to 0$ are dominated by noise, the aggregated global representation exhibits substantial uncertainty. Unconstrained integration of this  global context across the ODE trajectory risks propagating noise into individual local features, thereby degrading trajectory predictions.

To mitigate this, we introduce a novel  Dynamic Cross-Interaction (DCI) mechanism that temporally modulates the global-to-local feature conditioning. The modulation weight $g(t)$ is defined as a function of the integration progress $t$:
\begin{equation}
    g(t) = w_{\min} + (w_{\max} - w_{\min}) t^\gamma.
\end{equation}
Here, the boundary values $w_{\min}$ and $w_{\max}$ define the minimum and maximum conditioning strengths, respectively, while the hyperparameter $\gamma$ governs the non-linear profile of the transition. By bounding $g(0) = w_{\min}$, we suppresses noisy social messages during the initial stages, constraining the feature update to local individual dynamics. As $t \to 1$, $g(t)$ monotonically approaches $w_{\max}$, adaptively restoring the global multi-person dependencies. The localized feature update at layer $l$ is formulated as:
\begin{equation}
    \tilde{\mathbf{H}}_{\mathrm{loc}}^{(l)} = \mathbf{H}_{\mathrm{loc}}^{(l)} + \nu\Big( \mathbf{H}_{\mathrm{loc}}^{(l)} \odot \big(1 + g(t)\mathbf{I}^{(l)}\big) + g(t)\mathbf{E}^{(l)} + \tau(\mathbf{D}_c) \Big),
\end{equation}
where $\mathbf{I}^{(l)}$ and $\mathbf{E}^{(l)}$ denote learned affine parameters derived from the global branch. This explicit time-conditioning ensures that the deterministic spatial prior ($\mathbf{D}_c$) remains unattenuated, while the uncertain global context is gated, effectively stabilizing the early integration steps.

\paragraph{Joint Refinement and Bidirectional Fusion.}
Complementary to the residual modeling of the motion branch, a parallel joint branch preserves skeletal integrity, initialized by embedding the coarse estimation $\hat{\mathbf{Y}}_c$. Following the identical architectural design of the coarse stage, this branch refines joint features $\mathbf{H}_J^{(l)}$ via intra- and inter-joint mappings, and integrates them with motion features $\mathbf{H}_M^{(l)}$ through a bidirectional gated exchange at each layer $l$. Specifically, $\mathbf{H}_J^{(l)}$ injects structural priors to constrain the flow integration, while $\mathbf{H}_M^{(l)}$ supplies dynamic multi-person contexts. This synchronized fusion ensures the generated sequences remain kinematically valid and contextually aware.

\paragraph{Future Motion Prediction.}
The motion features $\mathbf{H}_M^{(L)}$ and the joint features $\mathbf{H}_J^{(L)}$ from the final refinement layer are first concatenated and projected via an MLP $\mu$. This aggregated feature updates the coarse latent feature $\mathbf{Z}_c$ to form the refined latent representation $\mathbf{Z}_f$:
\begin{equation}
    \mathbf{Z}_f = \mu([\mathbf{H}_M^{(L)}, \mathbf{H}_J^{(L)}]) + \mathbf{Z}_c.
\end{equation} 
Leveraging the endpoint parameterization, $K$ parallel prediction heads $\rho_k$ directly decode distinct multi-modal residuals in a single step. Specifically, the $k$-th head predicts the clean spatial endpoint residual $\hat{\mathbf{R}}^{(1)}_k = \mathrm{iDCT}(\rho_k(\mathbf{Z}_f))$. This predicted residual is then anchored to the last observed motion $\mathbf{X}_{T_h}$ to generate the raw $K$-mode future motion $\{\mathbf{Y}_k\}_{k=1}^K$:
\begin{equation}
    \mathbf{Y}_k = \hat{\mathbf{R}}^{(1)}_k + \mathbf{X}_{T_h}.
\end{equation}  
To model global temporal correlations and mitigate behavioral discontinuities, each prediction is concatenated with the observed history $\mathbf{X}$. Then, a full-sequence temporal alignment, which utilizes temporal attention on the unified trajectory $[\mathbf{X}, \mathbf{Y}_k]$, refines the whole motion dynamics. Finally, the refined sequence is truncated to isolate the future horizon, yielding the final multi-person motion prediction $\{\mathbf{Y}^*_k\}_{k=1}^K$.


\begin{table*}[t]
\centering
\footnotesize
\resizebox{\textwidth}{!}{%
\begin{tabular}{@{}ll *{10}{c} @{}}
\toprule
\textbf{Metric} & \textbf{Method} &
\multicolumn{2}{c}{3DPW} &
\multicolumn{2}{c}{CMU-Syn} &
\multicolumn{2}{c}{AMASS/3DPW} &
\multicolumn{2}{c}{CMU-Syn/MuPoTS} &
\multicolumn{2}{c}{Mocap-UMPM} \\
\cmidrule(lr){3-4}\cmidrule(lr){5-6}\cmidrule(lr){7-8}\cmidrule(lr){9-10}\cmidrule(lr){11-12}
& &
\shortstack{Ori\\16f/14f} & \shortstack{RC\\16f/14f} &
\shortstack{2s\\/2s} & \shortstack{1s\\/1s} &
\shortstack{Ori\\16f/14f} & \shortstack{RC\\16f/14f} &
\shortstack{2s\\/2s} & \shortstack{1s\\/1s} &
\shortstack{Mix1\\6P} & \shortstack{Mix2\\10P} \\
\midrule
MPJPE & T2P\,\textsuperscript{\citeyear{jeong2024cvpr-multiagent}}       & 127.12 & 111.82 & 72.42 & 35.00 & 110.20 & 96.12 & 76.16 & 42.65 & 41.74 & 45.98 \\
      & CoMusion\,\textsuperscript{\citeyear{sun2024eccv-comusion}}  & 162.24 & 132.66 & 59.97 & 62.59 & 129.62 & 128.41 & 140.09 & 119.79 & 53.74 & 88.96 \\
      & JRT\,\textsuperscript{\citeyear{xu2023jointr}}       & 93.89 & 83.00 & 54.44 & 18.09 & 97.36 & 80.36 & 62.67 & 36.82 & 27.71 & 31.91 \\
      & TBIFormer\,\textsuperscript{\citeyear{peng2023tbiformer}} & 151.57 & 123.57 & 64.56 & 22.48 & 148.54 & 123.99 & 81.84 & 40.24 & 62.43 & 67.70 \\
      & EMPMP\,\textsuperscript{\citeyear{zheng2025empmp}}  & 90.57 & 72.80 & 50.99 & 25.04 & 87.72 & 70.91 & 59.10 & 33.19 & 26.92 & 28.48 \\
      & Ours      & \textbf{86.65} & \textbf{69.92} & \textbf{30.65} & \textbf{13.43} & \textbf{84.70} & \textbf{69.63} & \textbf{57.84} & \textbf{32.06} & \textbf{21.31} & \textbf{20.05}
 \\
\midrule
VIM   & T2P\,\textsuperscript{\citeyear{jeong2024cvpr-multiagent}}       & 65.47 & 57.88 & 41.30 & 25.31 & 57.47 & 50.64 & 43.38 & 25.92 & 27.83 & 31.12 \\
      & CoMusion\,\textsuperscript{\citeyear{sun2024eccv-comusion}}  & 76.18 & 60.22 & 31.98 & 30.45 & 63.07 & 59.10 & 72.70 & 54.75 & 29.56 & 44.70 \\
      & JRT\,\textsuperscript{\citeyear{xu2023jointr}}       & 51.27 & 44.40 & 32.44 & 13.98 & 53.40 & 42.86 & 35.84 & 23.85 & 19.50 & 22.66 \\
      & TBIFormer\,\textsuperscript{\citeyear{peng2023tbiformer}} & 75.25 & 62.44 & 37.04 & 15.89 & 73.84 & 62.27 & 46.45 & 26.20 & 38.63 & 42.33 \\
      & EMPMP\,\textsuperscript{\citeyear{zheng2025empmp}}  & 48.94 & 38.40 & 30.16 & 19.10 & 47.37 & 37.39 & 34.32 & 21.60 & 18.84 & 19.68 \\
      & Ours      & \textbf{47.03} & \textbf{37.26} & \textbf{19.20} & \textbf{10.32} & \textbf{45.93} & \textbf{37.08} & \textbf{33.39} & \textbf{20.69} & \textbf{14.59} & \textbf{13.82} \\
\midrule
JPE   & T2P\,\textsuperscript{\citeyear{jeong2024cvpr-multiagent}}       & 225.39 & 186.69 & 130.79 & 65.31 & 195.63 & 162.68 & 124.86 & 68.50 & 79.50 & 87.06 \\
      & CoMusion\,\textsuperscript{\citeyear{sun2024eccv-comusion}}  & 245.74 & 185.66 & 94.25 & 76.54 & 207.10 & 178.33 & 221.00 & 138.23 & 79.49 & 118.24 \\
      & JRT\,\textsuperscript{\citeyear{xu2023jointr}}       & 176.86 & 142.79 & 102.26 & 36.21 & 187.12 & 138.44 & 108.45 & 63.82 & 54.01 & 61.93 \\
      & TBIFormer\,\textsuperscript{\citeyear{peng2023tbiformer}} & 255.53 & 199.33 & 119.09 & 42.04 & 249.06 & 198.29 & 149.04 & 70.52 & 111.92 & 120.06 \\
      & EMPMP\,\textsuperscript{\citeyear{zheng2025empmp}}  & 170.31 & 120.68 & 91.89 & 47.13 & 163.72 & \textbf{119.29} & \textbf{100.42} & \textbf{54.43} & 51.89 & 53.35 \\
      & Ours      & \textbf{161.73} & \textbf{118.71} & \textbf{59.35} & \textbf{26.44} & \textbf{159.33} & 120.46 & 104.30 & 55.45 & \textbf{40.60} & \textbf{37.36} \\
\midrule
APE   & T2P\,\textsuperscript{\citeyear{jeong2024cvpr-multiagent}}       & 125.63 & 125.40 & 74.85 & 49.55 & 118.28 & 116.16 & 87.36 & 55.17 & 47.38 & 54.85 \\
      & CoMusion\,\textsuperscript{\citeyear{sun2024eccv-comusion}}  & 136.91 & 128.75 & 65.48 & 51.46 & 126.14 & 125.58 & 157.44 & 114.93 & 56.66 & 76.58 \\
      & JRT\,\textsuperscript{\citeyear{xu2023jointr}}       & 121.29 & 117.77 & 68.25 & 29.47 & 118.92 & 113.17 & 92.66 & 57.92 & 39.55 & 46.27 \\
      & TBIFormer\,\textsuperscript{\citeyear{peng2023tbiformer}} & 123.73 & 123.64 & 75.89 & 33.25 & 123.54 & 123.99 & 111.21 & 58.10 & 46.24 & 55.25 \\
      & EMPMP\,\textsuperscript{\citeyear{zheng2025empmp}}  & 111.11 & 102.95 & 63.67 & 38.61 & 107.64 & 100.03 & 84.23 & 51.10 & 38.41 & 41.26 \\
      & Ours      & \textbf{103.93} & \textbf{99.05} & \textbf{42.86} & \textbf{22.72} & \textbf{103.06} & \textbf{95.63} & \textbf{82.85} & \textbf{50.62} & \textbf{29.16} & \textbf{28.29} \\
\midrule
FDE   & T2P\,\textsuperscript{\citeyear{jeong2024cvpr-multiagent}}       & 169.96 & 129.23 & 99.33 & 39.70 & 147.80 & 110.41 & 90.14 & 45.71 & 59.17 & 63.84 \\
      & CoMusion\,\textsuperscript{\citeyear{sun2024eccv-comusion}}  & 206.37 & 139.99 & 70.29 & 62.39 & 165.42 & 132.47 & 181.09 & 128.40 & 60.33 & 94.97 \\
      & JRT\,\textsuperscript{\citeyear{xu2023jointr}}       & 124.21 & 93.54 & 74.18 & 22.35 & 144.27 & 93.00 & 73.55 & 46.22 & 38.90 & 43.28 \\
      & TBIFormer\,\textsuperscript{\citeyear{peng2023tbiformer}} & 215.23 & 158.17 & 90.97 & 27.27 & 212.73 & 158.25 & 114.99 & 50.18 & 99.15 & 105.08 \\
      & EMPMP\,\textsuperscript{\citeyear{zheng2025empmp}}  & 126.48 & \textbf{71.76} & 64.73 & 29.67 & 122.94 & \textbf{76.62} & \textbf{69.78} & \textbf{38.83} & 36.22 & 35.38 \\
      & Ours      & \textbf{117.50} & 74.50 & \textbf{40.34} & \textbf{15.72} & \textbf{122.43} & 80.76 & 74.01 & 39.35 & \textbf{28.52} & \textbf{25.15} \\
\bottomrule
\end{tabular}%
}
\caption{Quantitative comparisons. 16f/14f denotes 16 input frames and 14 predicted frames; 6P/10P indicates 6 or 10 persons. VIM is measured in cm, while JPE, MPJPE, APE, and FDE are measured in mm.}
\label{tab:main_results}
\end{table*}

\subsection{Training Objectives}
\label{sec:loss}

The model is trained by jointly optimizing the deterministic coarse prior and residual flow refinement modules. First, an $L_2$ loss penalizes the deviation between the coarse prediction $\mathbf{Y}_c$ and ground truth $\mathbf{Y}^{GT}$. Second, our endpoint parameterization simplifies the flow matching objective into direct target regression. To capture multi-modal futures, we instantiate this as a Winner-Takes-All (WTA) objective, minimizing the $L_2$ error of the best-matching hypothesis among $K$ predictions $\{\mathbf{Y}^*_k\}_{k=1}^K$. Finally, geometric constraints penalize bone-length violations to ensure anatomical plausibility.

\section{Experiments}
\label{sec:experiments}

\subsection{Datasets and Evaluation Metrics}
\label{sec:datasets_metrics}

\begin{table*}[t]
  \centering
  \setlength{\tabcolsep}{3pt}
  \renewcommand{\arraystretch}{1.08}
  \resizebox{\textwidth}{!}{%
  \begin{tabular}{@{}l*{15}{c}@{}}
    \toprule
    \textbf{Metric}
      & \multicolumn{3}{c}{\textbf{MPJPE}}
      & \multicolumn{3}{c}{\textbf{VIM}}
      & \multicolumn{3}{c}{\textbf{JPE}}
      & \multicolumn{3}{c}{\textbf{APE}}
      & \multicolumn{3}{c}{\textbf{FDE}} \\
    \cmidrule(lr){2-4}\cmidrule(lr){5-7}\cmidrule(lr){8-10}\cmidrule(lr){11-13}\cmidrule(l){14-16}
    \textbf{Out Length}
      & 1s & 2s & 3s
      & 1s & 2s & 3s
      & 1s & 2s & 3s
      & 1s & 2s & 3s
      & 1s & 2s & 3s \\
    \midrule
    T2P       & 42.94 & 80.75 & 116.93 & 36.69 & 65.19 & 92.97 & 80.12 & 151.37 & 222.67 & 53.90 & 76.93 & 87.81 & 55.97 & 123.26 & 193.93 \\
    CoMusion  & 90.05 & 113.71 & 139.96 & 46.51 & 68.08 & 90.11 & 109.32 & 162.66 & 218.92 & 71.88 & 89.31 & 101.06 & 87.48 & 135.58 & 189.42 \\
    JRT       & 30.35 & 58.36 & 83.51 & 28.41 & 49.32 & 68.78 & 59.22 & 108.74 & 157.27 & 47.45 & 73.96 & 88.91 & 36.29 & 76.81 & 121.88 \\
    TBIFormer & 36.48 & 67.12 & 96.36 & 30.83 & 54.63 & 77.20 & 66.69 & 124.59 & 182.29 & 51.38 & 76.50 & 88.11 & 44.41 & 95.43 & 151.42 \\
    EMPMP  & 30.41 & 56.37 & 79.89 & 26.95 & 46.67 & 65.11 & 56.62 & 103.26 & 149.10 & 45.64 & 69.48 & 82.05 & 34.39 & 73.23 & 116.25 \\
    Ours      & \textbf{23.68} & \textbf{46.16} & \textbf{68.00} & \textbf{21.64} & \textbf{40.15} & \textbf{58.53} & \textbf{45.47} & \textbf{88.30} & \textbf{134.02} & \textbf{36.71} & \textbf{59.69} & \textbf{73.04} & \textbf{27.61} & \textbf{62.20} & \textbf{104.55} \\
    \bottomrule
  \end{tabular}%
  }
  \caption{Performance comparison on \textit{CMU-Syn} in the 1s-input/3s-output setting on various metrics. Lower is better.}
    \label{tab:cmu-syn-1s3s}
\end{table*}

\subsubsection{Intra-Dataset Evaluation.}
Models are trained and evaluated on the same dataset, including:
(1) \textit{\textbf{3DPW}}~\cite{vonmarcard2018_3dpw}: A real-world, in-the-wild dataset, featuring two interacting subjects in outdoor and indoor environments. Models are evaluated under a 16-frame input and 14-frame output horizon using both original (3DPW-Ori) and camera-motion-compensated (3DPW-RC) coordinates~\cite{xu2023jointr,zheng2025empmp}.
(2) \textit{\textbf{CMU-Syn}}~\cite{wang2021mrt}: A synthetic dataset constructed by blending motion capture sequences into 3-person scenes. Models are evaluated across short-term (1s/1s), long-term (2s/2s), and extended (1s/3s) forecasting horizons.
(3) \textit{\textbf{Mix1 \& Mix2}}: Following~\cite{peng2023tbiformer}, we use high-density benchmarks (6 and 10 persons) synthesized from CMU-Mocap, UMPM, 3DPW, and MuPoTS-3D for the 2s-input/1s-output setting. Additionally, we evaluate 2s/2s variants (50 frames for both input and output). Unlike original test-only usage, we establish custom train/test splits for evaluation.

\subsubsection{Cross-Dataset Evaluation.}
To assess transferability and the impact of large-scale priors, we use two settings:
(1) \textit{\textbf{AMASS}}~\cite{mahmood2019amass}: AMASS is a large-scale motion capture database containing highly diverse human motions. Models are pre-trained on AMASS and fine-tuned on 3DPW to evaluate the benefits of large-scale pre-training.
(2) \textit{\textbf{MuPoTS-3D}}~\cite{mehta2018mupots}: It consists of real-world, multi-person test sequences captured in  indoor and outdoor settings. Models trained on CMU-Syn are directly evaluated on MuPoTS-3D to test cross-dataset transferability~\cite{wang2021mrt,xu2023jointr,zheng2025empmp}.

\subsubsection{Evaluation Metrics.}
We evaluate predictions using five metrics: \textit{\textbf{MPJPE}} averages joint position errors across all frames; \textit{\textbf{JPE}} measures absolute global joint errors at specific timesteps; \textit{\textbf{APE}} evaluates local articulation errors by excluding root displacement; \textit{\textbf{FDE}} quantifies global trajectory deviations via hip translation errors; and \textit{\textbf{VIM}} computes root-mean-square pose errors across agents. All metrics are reported in millimeters (mm) except VIM (cm). 

More details on the experimental settings, dataset  and metrics are provided in the supplementary material.

\subsection{Experimental Results}

\paragraph{Baselines.}
We compare against recent state-of-the-art baselines in multi-person motion forecasting, including JRT~\cite{xu2023jointr}, TBIFormer~\cite{peng2023tbiformer}, T2P~\cite{jeong2024cvpr-multiagent}, and EMPMP~\cite{zheng2025empmp}, as well as the single-person stochastic
motion prediction method CoMusion~\cite{sun2024eccv-comusion}. 

\paragraph{Performance on Intra-Dataset Forecasting.}
Under the intra-dataset setting, our approach achieves the overall best performance across the synthetic CMU-Syn and real-world 3DPW benchmarks (Table~\ref{tab:main_results}). The overall superiority on 3DPW demonstrates the model's robustness in handling diverse in-the-wild scenarios. Specifically, the reductions in MPJPE and JPE indicate highly accurate overall
spatial alignments and global joint-position predictions at the
selected timesteps. The substantial improvements in APE under the camera-motion-compensated
(3DPW-RC) setting—which removes estimated camera motion—indicate  that
the framework successfully captures fine-grained local body
articulations, rather than merely fitting macroscopic global
trajectories (FDE). Furthermore, evaluations across varying temporal horizons on CMU-Syn, including short-term (1s/1s), long-term (2s/2s), and extended (1s/3s) settings (Table~\ref{tab:cmu-syn-1s3s}), show improved performance margins over baselines. These performances across prolonged sequences demonstrates that our model effectively mitigates the error accumulation  inherent in long-term forecasting.

To evaluate the model's capacity to handle highly crowded environments, we assess its performance on the Mix1 (6-person) and Mix2 (10-person) benchmarks. As detailed in Table~\ref{tab:main_results}, our method achieves the lowest errors across all five metrics on both datasets. Notably, the significant reductions in JPE (40.60 on Mix1, 37.36 on Mix2) and FDE (28.52 on Mix1, 25.15 on Mix2) demonstrate that the model accurately predicts specific individual poses and global root-trajectory predictions even when spatial constraints are severe. Crucially, the sustained accuracy in the 10-person setting indicates that the framework does not degrade under increased social complexity. Instead, the proposed interaction modeling effectively processes the expanded relational dynamics, leveraging dense social contexts to maintain high predictive accuracy and structural consistency.

\paragraph{Performance on Cross-Dataset Evaluation and Pre-training.}
To assess the model's robustness across varying data distributions and its capacity to benefit from large-scale data, we evaluate the framework under synthetic-to-real (CMU-Syn to MuPoTS-3D) and pre-training scale-up (AMASS to 3DPW) settings. When trained exclusively on synthetic CMU-Syn data and evaluated directly on real-world MuPoTS-3D sequences, our model achieves the lowest errors across primary spatial (MPJPE), pose-vector (VIM), and local articulation (APE) metrics. Furthermore, incorporating pre-training on the extensive AMASS database consistently improves the base model's performance on 3DPW across these primary metrics. Under this setting, the framework maintains state-of-the-art results in MPJPE, VIM, and APE, while yielding highly competitive global root-trajectory predictions (FDE). These  results suggest that the framework learns generalizable  representations rather than overfitting to the source domain, demonstrating its reliability in modeling multi-agent dynamics in unseen scenarios.

\subsection{Ablation Study}
\label{sec:ablation}

We conduct ablation studies on the 3DPW-RC and CMU-Syn (1s/1s) benchmarks to evaluate the contribution of each architectural design, with results  summarized in Table~\ref{tab:ablation}.

\begin{table}[t]
\centering
\small
\setlength{\tabcolsep}{4pt}
\begin{tabular}{@{}l|ccc|ccc@{}}
\toprule
& \multicolumn{3}{c|}{\textbf{3DPW-RC}} & \multicolumn{3}{c}{\textbf{CMU-Syn}} \\
\cmidrule(lr){2-4} \cmidrule(lr){5-7}
\textbf{Variant} & \textbf{VIM} & \textbf{APE} & \textbf{MPJPE} & \textbf{VIM} & \textbf{APE} & \textbf{MPJPE} \\
\midrule
Full model           & \textbf{37.26} & 99.05          & \textbf{69.92} & \textbf{10.32} & \textbf{22.72} & \textbf{13.43} \\
w/o DCP              & 38.76          & 101.62         & 73.30          & 11.17            & 24.50            & 14.54 \\
w/o RFM              & 38.95          & 101.08         & 72.70          & 11.77          & 25.32          & 15.25 \\
w/ AFM         & 38.85          & 102.17         & 73.48                & 12.24          & 26.07          & 16.35 \\
w/o DCI              & 37.50          & 100.55         & 70.62          & 11.12          & 24.55          & 14.43 \\
w/o JB               & 38.71          & 102.56         & 73.29          & 11.42          & 25.11          & 15.01 \\
w/o FTA   & 37.56      & \textbf{98.16} & 70.41                    & 10.66            & 23.29            & 13.88 \\

\bottomrule
\end{tabular}
\caption{Ablation study results on \textit{3DPW-RC} and \textit{CMU-Syn}. }
\label{tab:ablation}
\end{table}

\paragraph{Effectiveness  of Deterministic Coarse Prior.}
The \textit{w/o DCP} variant removes the deterministic coarse prior module, directly synthesizing multi-person motion sequences from historical observations with the flow matching module. This variant degrades the performance on both of the datasets, validating the effectiveness of extracting basic deterministic priors to guide the subsequent generative process.

\paragraph{Effectiveness  of Residual Flow Matching.}
Replacing the ODE integration with a deterministic network (\textit{w/o RFM}) increases errors across all metrics (MPJPE +2.78 on 3DPW-RC and +1.82 on CMU-Syn), confirming that flow-based refinement better captures fine-grained details. Furthermore, formulating the flow over absolute coordinates instead of motion residuals (\textit{w/ AFM}) also degrades performance. This indicates that predicting residuals provides a more stable optimization target than directly regressing absolute positions.

\paragraph{Effectiveness  of Dynamic Cross-Interaction.}
The DCI module utilizes a time-varying modulation weight $g(t)$ to regulate spatial message-passing during the iterative generation process. Replacing this dynamic mechanism with static Individual Social Interaction used in the coarse stage (\textit{w/o DCI}, i.e., $g(t)$ =1) leads to a clear degradation in overall performance. This proves that scaling  interaction strength along ODE trajectory is crucial for accurate motion generation.

\paragraph{Effectiveness  of Joint Branch.}
Removing the joint branch across both the coarse and refinement stages (\textit{w/o JB}) increases position errors. This confirms that explicit spatial feature extraction is crucial for accurate 3D joint motion prediction, effectively complementing the motion branch.

\paragraph{Effectiveness  of Full-Sequence Temporal Alignment.}
Removing the global attention across concatenated sequences (\textit{w/o FTA}) increases the overall prediction errors on both datasets. This confirms that full-sequence temporal modeling is essential for mitigating behavioral discontinuities between historical observations and future predictions.

More ablation studies on the hyper-parameters, loss function, and visualizations are in the supplementary material.

\subsection{Qualitative Results}
\label{sec:qualitative}

\begin{figure}[!t]
    \centering
    \includegraphics[width=1.0\linewidth]{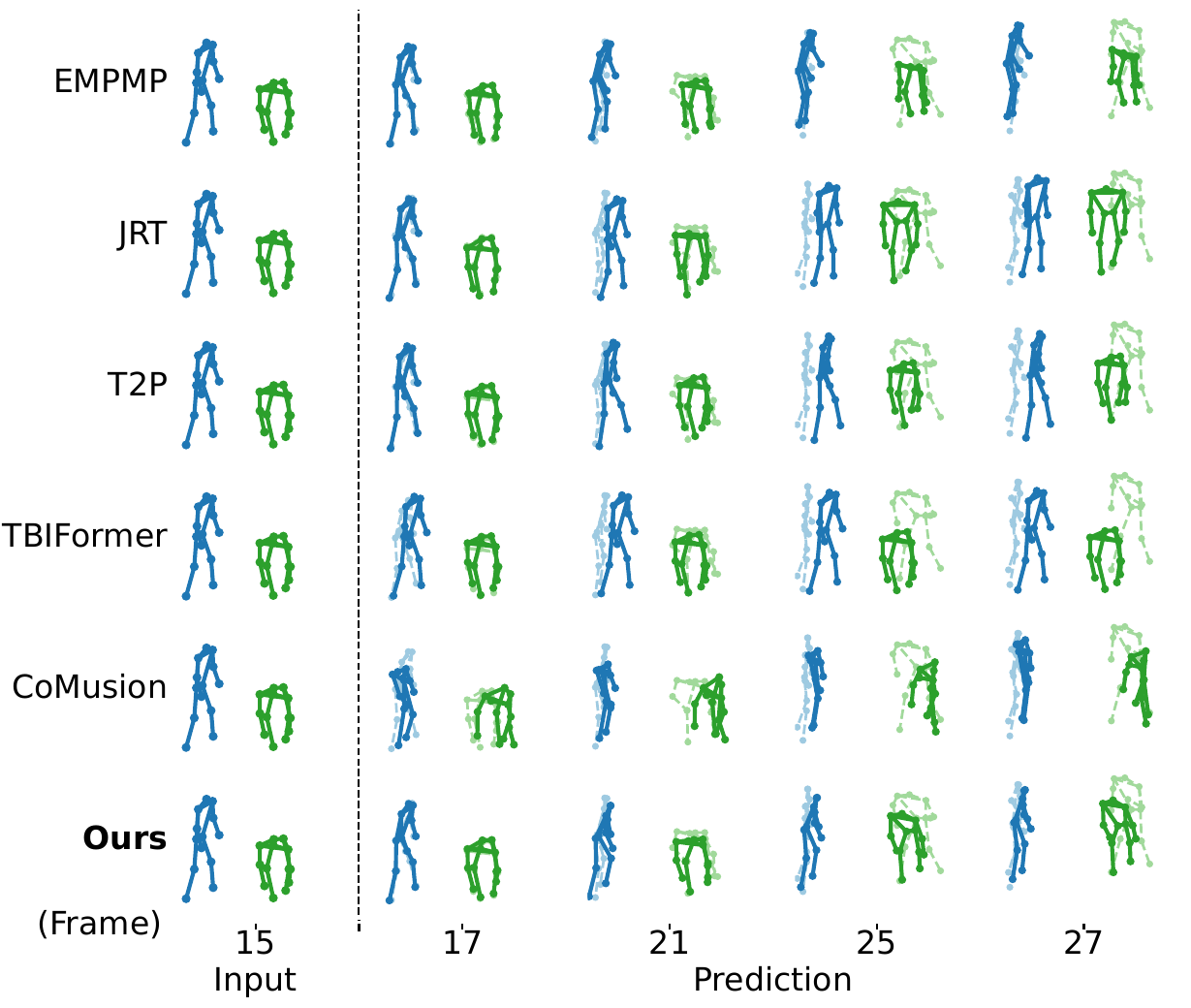}
    \setlength{\abovecaptionskip}{-2pt}
    \setlength{\belowcaptionskip}{-6pt}
    \caption{Visualization on  \textit{3DPW-RC}. Solid lines denote predictions and faded lines represent ground truth. Our model captures the subject's large-scale bend-to-stand motion while maintaining  spatial alignment for both interacting subjects.}
    \label{fig:3dpw_picture1}
\end{figure}

We evaluate a challenging sequence from \textit{3DPW-RC} where the green subject performs a large-scale bend-to-stand motion while the blue subject remains relatively stable. As shown in Fig.~\ref{fig:3dpw_picture1}, EMPMP, T2P, and TBIFormer struggle to capture this drastic posture change, erroneously keeping the green subject bent over in later frames. JRT anticipates the standing motion but suffers from severe spatial misalignment. While CoMusion is competitive, our model most accurately tracks the green subject's upward recovery and maintains the blue subject's stability, aligning closely with the ground truth (faded). This confirms that our approach effectively captures complex motion intentions and preserves precise spatial coherence during asymmetric interactions.

\section{Conclusion}
We present a Prior-Guided Residual Flow Matching framework for 3D multi-person motion prediction. It first extracts a Deterministic Coarse Prior (DCP) to anchor a Residual Flow Matching (RFM) module, which formulates the continuous-time flow over motion residuals while preserving skeletal integrity via bidirectional joint-motion fusion. A Dynamic Cross-Interaction (DCI) mechanism temporally modulates inter-agent message passing to mitigate early-stage ODE noise interference. Experiments demonstrate state-of-the-art performance across diverse benchmarks on multiple metrics.

Despite its effectiveness, the proposed framework presents specific limitations. First, the iterative ODE integration in the flow matching process incurs higher inference latency. Second, while the Dynamic Cross-Interaction mechanism stabilizes early-stage generation, its temporal schedule relies on predefined hyperparameters and does not yet incorporate external scene constraints. Third, following the spatial modeling paradigm of EMPMP, the current architecture assumes a fixed number of interacting individuals, restricting its flexibility. Future work will explore distillation techniques to accelerate inference and investigate scene-aware modulation  to ground human interactions within physical environments.

\bibliography{aaai2027}


\clearpage
\appendix
\twocolumn[
\begin{center}
{\LARGE\bfseries Residual Flow Matching with Dynamic Cross-Interaction for 3D Multi-Person Motion Prediction \\
\vspace{1.0em}
\Large Supplementary Material}
\end{center}
\vspace{1.2em}
]
\setlength{\parindent}{0pt}

\subsection{Details on Dataset}

\label{sec:appendix_datasets}

We evaluate our framework on 3DPW~\cite{vonmarcard2018_3dpw}, AMASS~\cite{mahmood2019amass}, MuPoTS-3D~\cite{mehta2018mupots}, as well as synthesized multi-person benchmarks derived from CMU motion capture sequences~\cite{wang2021mrt,peng2023tbiformer}. All of these datasets are publicly available. These datasets encompass both in-the-wild and synthetic environments, enabling a comprehensive assessment of the model's accuracy, scalability, and cross-domain generalization.

\paragraph{3DPW.}
3DPW~\cite{vonmarcard2018_3dpw} is a real-world in-the-wild dataset captured by moving cameras and IMU-assisted body reconstructions. It contains challenging interactions, frequent camera motion, partial occlusions, and diverse indoor and outdoor scenes. Following the standard protocol ~\cite{vendrow2022somoformer,xu2023jointr,zheng2025empmp}, we evaluate both 3DPW-Ori and 3DPW-RC. The former uses the original coordinates, while the latter removes shared camera motion. The forecasting horizon is set to 16 observed frames and 14 predicted frames. Training uses {24} selected 3DPW scene files, whereas evaluation uses {170} pre-packaged two-person clips.

\paragraph{AMASS.}
AMASS~\cite{mahmood2019amass} is a large-scale motion-capture corpus providing a unified representation across multiple public datasets. In our experiments, it serves as the source domain to learn generalized motion priors prior to fine-tuning on 3DPW-Ori and 3DPW-RC. Specifically, we utilize 1,983 motion sequences from the CMU subset for pre-training. Following temporal downsampling, sliding-window segmentation, and pairing, these sequences yield 108,583 two-person training clips. To maintain representational consistency, the coordinate normalization is strictly aligned with the downstream 3DPW datasets for the AMASS/3DPW-Ori and AMASS/3DPW-RC settings.

\paragraph{CMU-Syn.}
CMU-Syn~\cite{wang2021mrt} is a synthetic dataset constructed by blending individual CMU motion capture sequences into multi-person interactive scenarios. It provides controllable crowd sizes and consistent protocol definitions, making it ideal for systematically comparing trajectory-conditioned and interaction-aware models over extended horizons. We evaluate the model across three temporal settings: short-term (1s/1s), long-term (2s/2s), and extended (1s/3s). The dataset comprises about {13000} sequences for training and {3000} sequences for testing.

\paragraph{MuPoTS-3D.}
MuPoTS-3D~\cite{mehta2018mupots}  is a real-world multi-person dataset characterized by complex interactions, severe occlusions, and dynamic viewpoint changes.  We utilize this dataset to evaluate zero-shot cross-dataset transferability. Specifically, models trained solely on the synthetic CMU-Syn dataset are directly tested on MuPoTS-3D to assess their robustness against domain shifts and real-world crowded scenes.  The evaluation is conducted under 1s/1s and 2s/2s forecasting protocols using {192} preprocessed three-person clips derived from the {20} original test sequences, with pretrained model on the CMU-Syn dataset.

\paragraph{Mix1 \& Mix2.}
~\cite{wang2021mrt,peng2023tbiformer} are high-density synthetic benchmarks containing 6 and 10 interacting individuals, respectively. These datasets are utilized to evaluate the model's performance in complex and crowded multi-person scenarios. For the 2,s/1,s setting, we configure both Mix1 and Mix2 to include 800 training and 100 testing sequences, with models trained and evaluated directly on these splits.
Furthermore, we construct a 2,s/2,s version of the Mix1 and Mix2 datasets, following the multi-person synthesis procedure of TBIFormer~\cite{peng2023tbiformer}. For both of the datasets, we synthesize 800 training and 100 testing scenes. The motion sequences are resampled to 25 FPS, resulting in 100-frame sequences (50 input and 50 target frames).

To comprehensively assess the proposed framework, we conduct evaluations under both intra-dataset and cross-dataset configurations. Intra-dataset evaluations are performed on 3DPW (including both 3DPW-Ori and 3DPW-RC), CMU-Syn, and the high-density Mix1 and Mix2 benchmarks, where models are trained and tested within the same domain. For cross-dataset and transfer learning settings, we evaluate the zero-shot generalization of models optimized on CMU-Syn when directly applied to MuPoTS-3D. Additionally, we investigate the efficacy of large-scale prior learning by fine-tuning AMASS-pretrained models on 3DPW. To ensure fair comparisons, all baseline methods are reproduced using the official codes under identical experimental protocols.

\subsection{Details on Metrics}

Formally, we denote the predicted future motion sequence as $\mathbf{Y}=\{\mathbf{y}_{p,j,t}\}\in\mathbb{R}^{ P\times T_f\times J\times 3}$, where $P$ represents the number of interacting individuals, $T'$ denotes the number of future frames, and $J$ indicates the number of joints per person. The corresponding ground truth motion is defined as $\mathbf{Y}^{GT}=\{\mathbf{y}_{p,j,t}^{GT}\}\in\mathbb{R}^{P\times T_f\times J\times 3}$. Here, $\mathbf{y}_{p,j,t}\in\mathbb{R}^3$ represents the 3D spatial coordinate of joint $j$ for person $p$ at future frame $t$. The  hip joint is set as the kinematic root.

\paragraph{MPJPE.}
The Mean Per Joint Position Error (MPJPE) quantifies the overall joint prediction accuracy by calculating the average Euclidean distance across all valid persons, joints, and predicted frames up to the reporting horizon:
\begin{equation}
    \mathrm{MPJPE}
=
\frac{1}{P T_f J}
\sum_{p=1}^{P}\sum_{t=1}^{T_f}\sum_{j=1}^{J}
\left\|\mathbf{y}_{p,j,t}-\mathbf{y}_{p,j,t}^{GT}\right\|_2.
\end{equation}
This metric is used to measure the overall spatial accuracy of the predicted motions across the forecasting window.

\paragraph{JPE.}
The Joint Position Error (JPE) evaluates the absolute joint localization accuracy at a specific future horizon $t$:
\begin{equation}
    \mathrm{JPE}@t
=
\frac{1}{PJ}
\sum_{p=1}^{P}\sum_{j=1}^{J}
\left\|\mathbf{y}_{p,j,t}-\mathbf{y}_{p,j,t}^{GT}\right\|_2.
\end{equation}
Unlike MPJPE, JPE measures the prediction error at a single specific frame.

\paragraph{APE.}
The Aligned Pose Error (APE) assesses the pure local articulation fidelity by decoupling the global root displacement of each individual:
\begin{equation}
\begin{split}
    &\mathrm{APE}@t
=\\
&\frac{1}{PJ}
\sum_{p=1}^{P}\sum_{j=1}^{J}
\left\|
\bigl(\mathbf{y}_{p,j,t}-\mathbf{y}_{p,\mathrm{hip},t}\bigr)-
\bigl(\mathbf{y}_{p,j,t}^{GT}-\mathbf{y}_{p,\mathrm{hip},t}^{GT}\bigr)
\right\|_2.
\end{split}
\end{equation}
It assesses the local articulation fidelity by decoupling the global root displacement.

\paragraph{FDE.}
The Final Displacement Error (FDE) evaluates the global trajectory displacement by measuring the root-joint deviation at a specific future horizon $t$:
\begin{equation}
    \mathrm{FDE}@t
=
\frac{1}{P}
\sum_{p=1}^{P}
\left\|\mathbf{y}_{p,\mathrm{hip},t}-\mathbf{y}_{p,\mathrm{hip},t}^{GT}\right\|_2.
\end{equation}
This metric focuses  on the accuracy of the global trajectory.

\paragraph{VIM.}
The Visibility-Ignored Metric (VIM) focuses on the full-body pose-vector error for a specific time frame. The VIM score at future horizon $t$ is calculated by accumulating the squared joint errors for each person, taking the square root, and then averaging across all valid individuals:
\begin{equation}
    \mathrm{VIM}@t
=
\frac{1}{P}
\sum_{p=1}^{P}
\sqrt{
\sum_{j=1}^{J}
\left\|\mathbf{y}_{p,j,t}-\mathbf{y}_{p,j,t}^{GT}\right\|_2^2
}.
\end{equation}
This metric provides a holistic evaluation of the full-body pose structure at specific frames.

\paragraph{Reporting Horizons and Scaling.}
The specific evaluation horizons for each metric are detailed in Table~\ref{tab:metric_frames}. The evaluation protocols for 3DPW, CMU-Syn, and MuPoTS-3D strictly follow EMPMP~\cite{zheng2025empmp}, with a coordinate conversion factor of 1/1.8 applied to CMU-Syn and MuPoTS-3D prior to error computation. For Mix1 and Mix2, we use a custom protocol for long-term evaluation. VIM is scaled by 
100 (centimeters, cm), while MPJPE, JPE, APE, and FDE are scaled by 1000 (millimeters, mm). Lower values indicate better performance across all metrics.

\begin{table}[h]
\centering
\small
\caption{Evaluation horizons for each benchmark. All frame indices are zero-based. MPJPE indices denote the endpoints of cumulative prediction prefixes.}
\label{tab:metric_frames}
\resizebox{\linewidth}{!}{%
\begin{tabular}{@{}lccc@{}}
\toprule
Protocol & VIM indices & MPJPE horizon indices & JPE / APE / FDE indices \\
\midrule
3DPW
& 1, 3, 7, 9, 13
& 6, 13
& 6, 13 \\
CMU-Syn / MuPoTS-3D 1s/1s
& 1, 3, 7, 11, 14
& 4, 9, 14
& 4, 9, 14 \\
CMU-Syn / MuPoTS-3D 2s/2s
& 1, 5, 10, 20, 29
& 9, 19, 29
& 9, 19, 29 \\
Mix1 \& Mix2 2s/1s
& 1, 5, 10, 15, 24
& 4, 9, 14, 19, 24
& 4, 9, 14, 19, 24 \\
Mix1 \& Mix2 2s/2s
& 9, 19, 29, 39, 49
& 9, 19, 29, 39, 49
& 9, 19, 29, 39, 49 \\
CMU-Syn 1s/3s
& 14, 29, 44
& 14, 29, 44
& 14, 29, 44 \\
\bottomrule
\end{tabular}%
}
\end{table}

\subsection{Details on Network and Training}

\paragraph{Network Architecture.} 
The model contains  the Deterministic Coarse Prior (DCP) stage and the Prior-Guided Residual Flow Refinement stage. The DCP stage comprises 16 encoding layers and 1 cross-branch interaction module, while the refinement stage contains 32 refinement layers and 3 cross-branch interaction modules. These cross-branch modules serve to synchronize the spatial joint kinematics and temporal motion patterns. Additionally, within the motion branch, a global-local interaction mechanism is systematically interleaved every four encoding/refinement layers to integrate individual dynamics with global social contexts.

To model temporal dynamics across varying observation ($T_h$) and prediction ($T_f$) horizons, we apply a Discrete Cosine Transform (DCT) along the temporal axis. To accommodate asymmetric lengths, we define a unified sequence length $I = \max(T_h, T_f)$. For scenarios where the prediction horizon exceeds the observation (e.g., $T_h < T_f$ in the 1s/3s setting), the historical sequence is first padded to length $I$ by replicating the last observed pose before applying the $I$-point DCT. The prediction heads then output $I$ frequency coefficients representing the future motion residuals. These coefficients are mapped back to the spatial domain via an $I$-point inverse DCT (iDCT). Finally, to  handle cases where the observation is longer than the prediction (e.g., $T_h > T_f$), the reconstructed spatial sequence is simply truncated to the first $T_f$ frames to yield the  prediction.



Within the refinement stage, The noise scale $\sigma$ used for flow state  initialization is set as 2.0. Within the Dynamic Cross-Interaction (DCI), the time-varying modulation weight is instantiated as $g(t) = 0.6 + 0.4t^{1.5}$, corresponding to a minimum conditioning strength $w_{\min} = 0.6$, a maximum strength $w_{\max} = 1.0$, and a non-linear transition profile $\gamma = 1.5$. Additionally, the global contribution weight $\alpha$, which controls the fusion of local individual features and global social states, is fixed at $0.2$ in the DCP stage. For the final motion prediction, the full-sequence temporal alignment module is implemented as a Transformer utilizing a single encoder layer, three attention heads, zero dropout, and a feed-forward dimension twice the hidden width. The hidden dimension is set to 39 for 3DPW and 45 for the CMU-Syn, MuPoTS-3D, and Mix benchmarks.

Finally, all multi-hypothesis  evaluations strictly adhere to the standard Best-of-$K$ protocol~\cite{jeong2024cvpr-multiagent} with $K=6$.

\paragraph{Training and Optimization}
The framework is optimized end-to-end using the Adam optimizer with a weight decay of $10^{-4}$. The model employs a batch size of 64, with gradient accumulation set to 1 and automatic mixed precision (AMP) disabled to ensure numerical stability during the ODE integration.

For AMASS pretraining and subsequent 3DPW fine-tuning, the model is trained for 50,000 updates with a constant learning rate of $3\times10^{-4}$. To prevent optimization bias, the optimizer state is strictly reinitialized prior to the fine-tuning phase. Conversely, for the CMU-Syn, MuPoTS-3D, and Mix benchmarks, the 50,000-update training regime utilizes a dynamic learning rate schedule: a 500-update linear warm-up to $3\times10^{-4}$, followed by a cosine decay to $10^{-6}$. 

During training, we apply two augmentations jointly to the observed and future sequences. First, each scene is rotated around the world vertical axis by an angle sampled uniformly from $[0,2\pi)$. Second, we randomly permute the ordering of persons using the same permutation for both the input and target sequences. 

The complete source code and trained models of our method will be fully released to the public once this paper is accepted.

\subsection{Flow Construction and ODE Sampling}

During training, the integration time step is sampled uniformly, $t \sim \mathcal{U}(0,1)$. Given the ground-truth target motion residual $\mathbf{R}^{(1)}$ and an initial noise sample $\mathbf{R}^{(0)} \sim \mathcal{N}(\mathbf{0}, \mathbf{I})$, the intermediate flow state $\mathbf{R}^{(t)}$ is constructed via linear interpolation:
\begin{equation} 
\mathbf{R}^{(t)} = t \mathbf{R}^{(1)} + (1-t) \sigma \mathbf{R}^{(0)},  
\end{equation}
where the noise scale is set as $\sigma = 2.0$.

At inference, motion generation is formulated as solving the induced Ordinary Differential Equation (ODE). We employ a fixed-step explicit Euler solver with 10 integration steps over the interval $t \in [0, 1)$. Given a constant step size of $\Delta t = 0.1$, the discrete state update is computed as:
\begin{equation}
    \mathbf{R}^{(t+\Delta t)} = \mathbf{R}^{(t)} + \frac{\hat{\mathbf{R}}^{(1)} - \mathbf{R}^{(t)}}{1 - t + 10^{-4}} \Delta t,
\end{equation}
where $\hat{\mathbf{R}}^{(1)}$ denotes the residual predicted by the network. 





\subsection{Additional Experimental Results}

\paragraph{Results on Mix1 \& Mix2 under 2s/2s Setting.}
Table~\ref{tab:mix_results} presents the quantitative evaluation on the Mix1 (6 person) and Mix2 (10 person) benchmarks under the 2s/2s setting. These datasets feature dense multi-person scenarios with social interactions, requiring models to effectively balance local kinematics and global dynamics. As shown in the table, our proposed framework achieves the overall best performance compared to the baseline methods.  Our method effectively handles individual motion modeling within complex multi-person scenes over extended temporal horizons.

\begin{table}[!t]
\centering
\small
\caption{Quantitative comparisons on the Mix1 and Mix2 benchmarks under the 2s/2s setting.}
\label{tab:mix_results}
\begin{tabular}{@{}llcc@{}}
\toprule
Metric & Method & Mix1 (2s/2s) & Mix2 (2s/2s) \\ \midrule
\multirow{6}{*}{MPJPE}
 & T2P$^{2024}$ & 177.0180 & 174.8503 \\
 & CoMusion$^{2024}$ & 191.6690 & 203.8496 \\
 & JRT$^{2023}$ & 114.2540 & 113.1434 \\
 & TBIformer$^{2023}$ & 181.4450 & 186.8490 \\
 & EMPMP$^{2025}$ & 116.9590 & 123.5478 \\
 & Ours & \textbf{107.9016} & \textbf{109.6569} \\ \midrule
\multirow{6}{*}{VIM}
 & T2P$^{2024}$ & 120.9714 & 122.5426 \\
 & CoMusion$^{2024}$ & 117.1240 & 124.1540 \\
 & JRT$^{2023}$ & 82.6251 & 83.9390 \\
 & TBIformer$^{2023}$ & 124.7099 & 131.0678 \\
 & EMPMP$^{2025}$ & 83.6671 & 89.0836 \\
 & Ours & \textbf{79.0911} & \textbf{79.9816} \\ \midrule
\multirow{6}{*}{JPE}
 & T2P$^{2024}$ & 298.0269 & 301.7580 \\
 & CoMusion$^{2024}$ & 290.6670 & 308.6260 \\
 & JRT$^{2023}$ & 197.4994 & 200.5367 \\
 & TBIformer$^{2023}$ & 305.8979 & 323.0130 \\
 & EMPMP$^{2025}$ & 200.1155 & 213.6713 \\
 & Ours & \textbf{191.5872} & \textbf{193.4759} \\ \midrule
\multirow{6}{*}{APE}
 & T2P$^{2024}$ & 102.2191 & 100.1616 \\
 & CoMusion$^{2024}$ & 130.7410 & 129.6372 \\
 & JRT$^{2023}$ & 95.0511 & 87.0158 \\
 & TBIformer$^{2023}$ & 100.2990 & 91.0160 \\
 & EMPMP$^{2025}$ & 87.7956 & 86.1416 \\
 & Ours & \textbf{82.0366} & \textbf{79.1080} \\ \midrule
\multirow{6}{*}{FDE}
 & T2P$^{2024}$ & 268.8525 & 275.4606 \\
 & CoMusion$^{2024}$ & 262.6870 & 277.9200 \\
 & JRT$^{2023}$ & \textbf{162.1737} & 167.6482 \\
 & TBIformer$^{2023}$ & 279.5892 & 300.3575 \\
 & EMPMP$^{2025}$ & 168.6042 & 183.8845 \\
 & Ours & 164.9555 & \textbf{166.0057} \\ \bottomrule
\end{tabular}
\end{table}

\paragraph{Ablation Study on Loss Function.}
We evaluate the contribution of specific training objectives in Table~\ref{tab:loss_ablation} by individually omitting the coarse loss ($\mathcal{L}_{\mathrm{coarse}}$) and the bone length constraint ($\mathcal{L}_{\mathrm{bone}}$) from the full formulation. Removing $\mathcal{L}_{\mathrm{coarse}}$ discards the explicit supervision for the Deterministic Coarse Prior (DCP), which compromises the stable kinematic foundation required for the subsequent flow generation. Furthermore, omitting $\mathcal{L}_{\mathrm{bone}}$ removes the penalty for anatomical violations during training, leading to a degradation in structural integrity. The full model achieves the best performance, demonstrating the necessity of combining deterministic structural guidance with  geometric constraints.

\begin{table}[h]
\centering
\small
\caption{Ablation study on the training objectives. Models are evaluated on 3DPW-RC and CMU-Syn (1s/1s) datasets.}
\label{tab:loss_ablation}
\begin{tabular}{@{}lcccccc@{}}
\toprule
Variant & \multicolumn{3}{c}{3DPW-RC} & \multicolumn{3}{c}{CMU-Syn} \\
\cmidrule(lr){2-4} \cmidrule(lr){5-7}
& VIM & APE & MPJPE & VIM & APE & MPJPE \\
\midrule
Full model & \textbf{37.26} & 99.05 & \textbf{69.92} & \textbf{10.32} & \textbf{22.72} & \textbf{13.43} \\
w/o $\mathcal{L}_{\mathrm{bone}}$ & 38.03 & \textbf{98.69} & 71.05 & 10.38 & 22.96 & 13.48 \\
w/o $\mathcal{L}_{\mathrm{coarse}}$ & 38.71 & 100.97 & 72.77 & 11.02 & 24.35 & 14.36 \\
\bottomrule
\end{tabular}
\end{table}

\begin{figure}[!tp]
    \centering
    \includegraphics[width=1.0\linewidth,height=6cm]{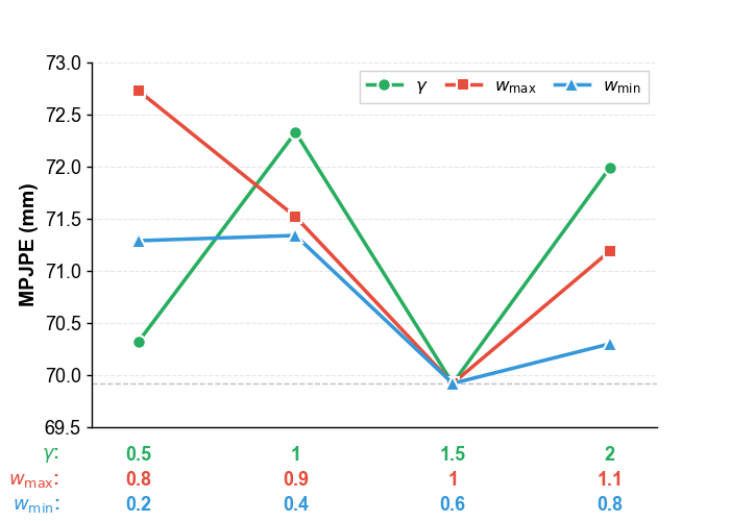}
    \setlength{\abovecaptionskip}{-2pt}
    \setlength{\belowcaptionskip}{-6pt}
    \caption{Performance curves of DCI hyperparameters evaluated by MPJPE. We analyze the impact of the transition profile $\gamma$ (green curve), the maximum conditioning strength $w_{\mathrm{max}}$ (red curve), and the minimum conditioning strength $w_{\mathrm{min}}$ (blue curve). The lowest prediction error is achieved at our default setting ($\gamma=1.5$, $w_{\mathrm{max}}=1.0$, $w_{\mathrm{min}}=0.6$).}
    \label{fig:hyperparameters}
\end{figure}

\paragraph{Ablation Study on DCI Hyperparameters.}
In the main text, we demonstrated the effectiveness of the Dynamic Cross-Interaction (DCI) mechanism compared to a static baseline. Here, we ablate the hyperparameters governing its time-varying modulation weight, defined as $g(t) = w_{\min} + (w_{\max} - w_{\min}) t^\gamma$. As shown in Figure~\ref{fig:hyperparameters}, we evaluate the maximum strength $w_{\max}$, the initial strength $w_{\min}$, and the transition profile $\gamma$.


We first analyze the maximum conditioning strength $w_{\max}$ (red curve), which controls the interaction as the latent state approaches the clean data distribution ($t \to 1$). Setting $w_{\max} = 1.0$ yields the lowest error. Lower values (e.g., $0.8$) restrict the model from fully utilizing multi-agent social contexts at the final integration stages, while exceeding this bound ($w_{\max} = 1.1$) slightly degrades performance, likely due to the over-amplification of global features.

Next, we evaluate the initial conditioning strength $w_{\min}$ (blue curve), which regulates the global-local interaction during the early, noise-dominated stages ($t \to 0$). A lower value ($w_{\min} \le 0.4$) results in higher errors, as it limits the integration of beneficial coarse spatial priors provided by the deterministic stage. Conversely, a higher initial weight ($w_{\min}=0.8$) also increases the error, as it allows early-stage noise from neighboring agents to interfere with local feature refinement. An intermediate setting of $w_{\min}=0.6$ provides a suitable balance.

Finally, we investigate the transition profile $\gamma$ (green curve). The proposed convex schedule ($\gamma=1.5$) achieves the best performance. As discussed in our method, this non-linear profile maintains noise suppression during early integration while progressively restoring multi-agent dependencies as trajectories clarify. A linear schedule ($\gamma=1.0$) or a more delayed schedule ($\gamma=2.0$) results in higher errors. While a concave schedule ($\gamma=0.5$) performs better than the linear one, it still falls short of $\gamma=1.5$, indicating that a moderate, convex delay in introducing global context is more effective for this formulation.

\subsection{More Visualizations}

\begin{figure*}[!tp]
    \centering
    \includegraphics[
        width=0.9\textwidth,
        keepaspectratio
    ]{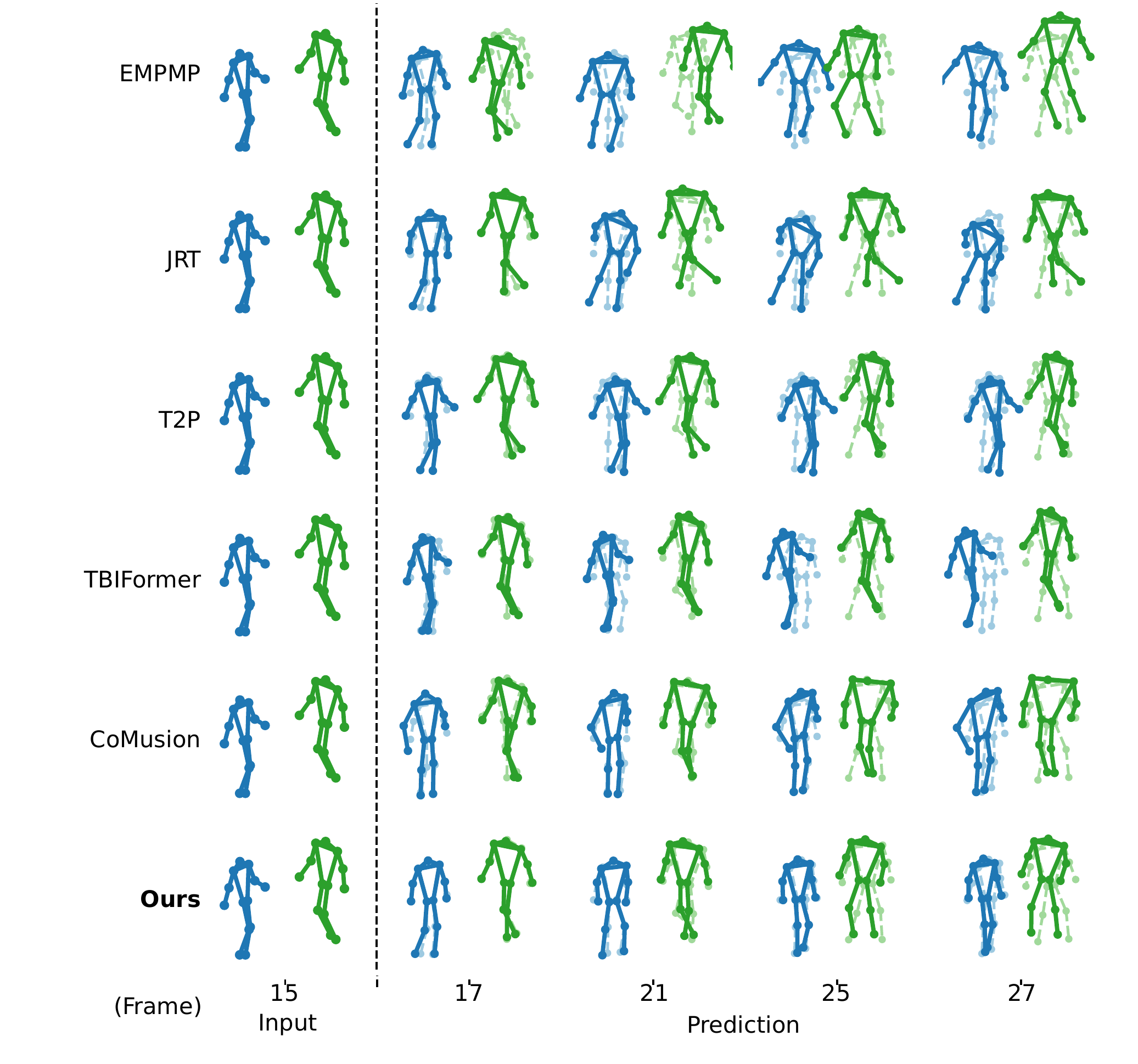}
    \setlength{\abovecaptionskip}{-2pt}
    \setlength{\belowcaptionskip}{-6pt}
    \caption{Visualization on \textit{3DPW-Ori}. Solid lines denote predictions and faded lines represent ground truth. Our model accurately preserves the relative positions and orientations between the two interacting subjects throughout the prediction sequence, maintaining consistent spatial relationships.}
    \label{fig:3dpw_ori_picture}
\end{figure*}

\begin{figure*}[!tp]
    \centering
    \includegraphics[
        width=\textwidth,
    ]{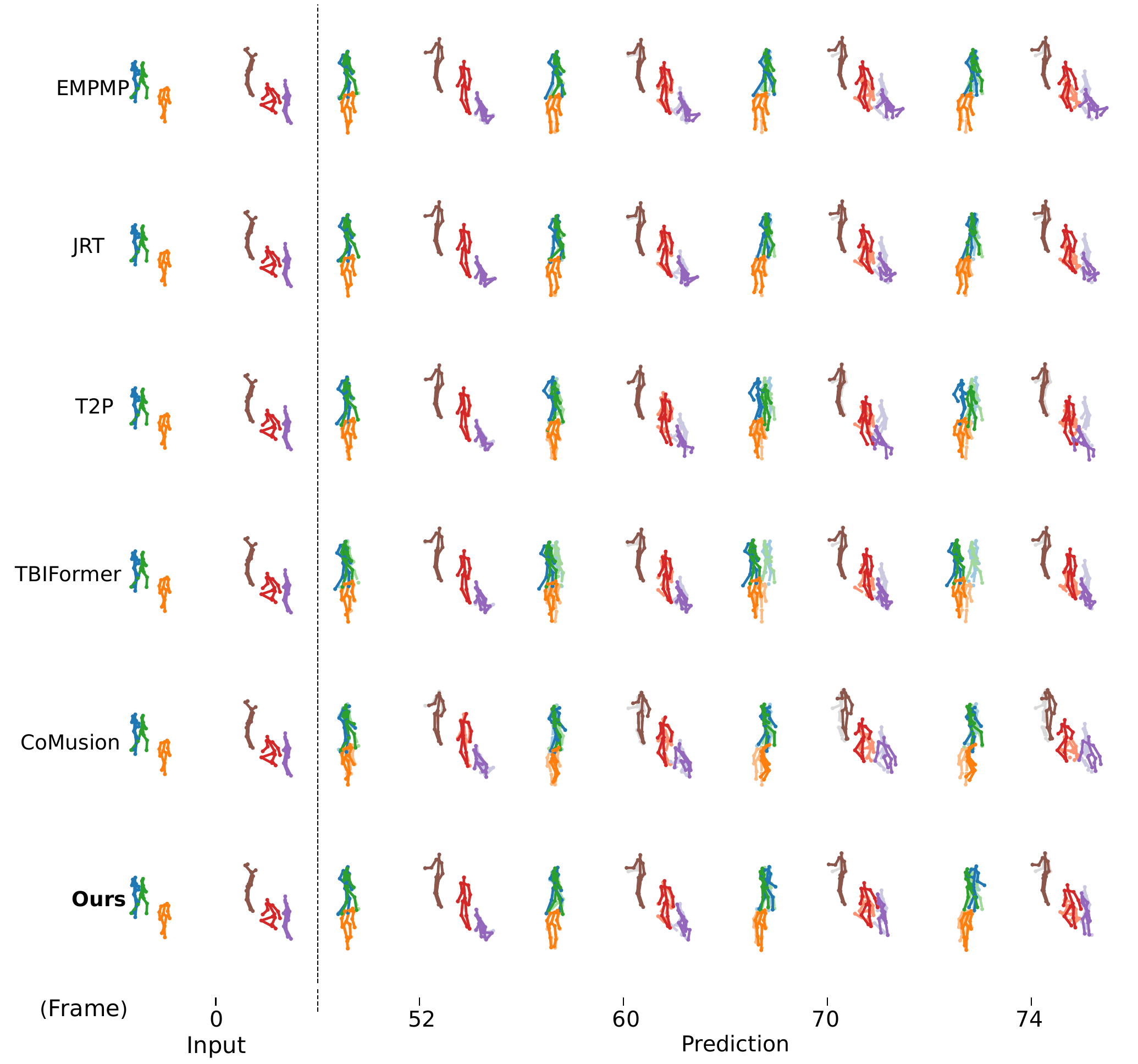}
    \caption{Additional qualitative comparisons on the Mix1 dataset under the 2s/1s setting. Light dashed skeletons denote the ground-truth future motions.}
    \label{fig:mix1_visualization}
\end{figure*}

\begin{figure*}[!tp]
    \centering
    \includegraphics[
        width=0.8\textwidth,
        height=0.8\textheight,
        keepaspectratio
    ]{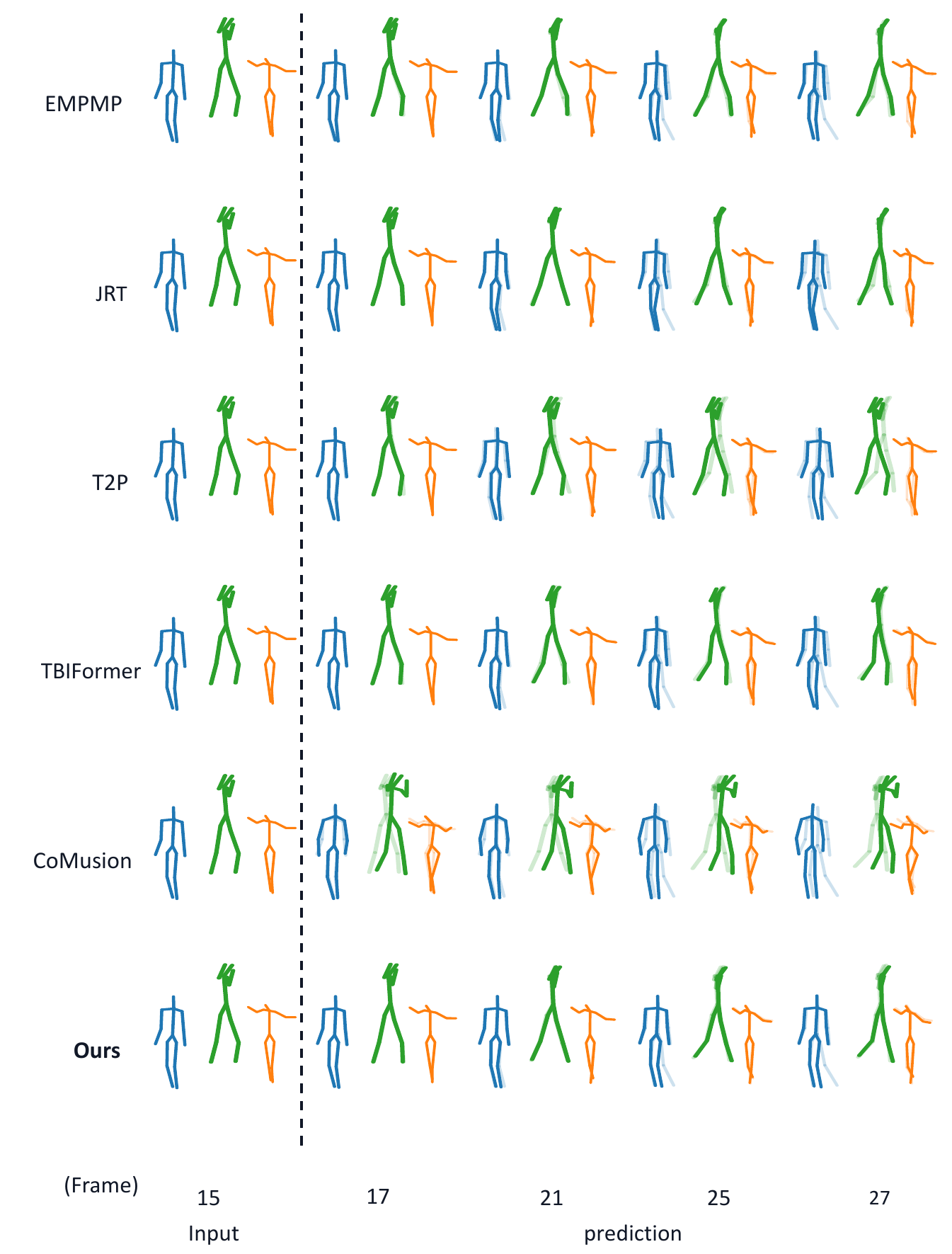}
    \setlength{\abovecaptionskip}{-2pt}
    \setlength{\belowcaptionskip}{-4pt}
    \caption{Qualitative comparison on the CMU-Syn dataset under the 1s/3s setting. Solid skeletons denote predicted motions, while faint skeletons indicate the ground-truth future motions.}
    \label{fig:cmu_syn_1s3s_04}
\end{figure*}

We provide additional qualitative comparisons against baseline methods across the Mix1, 3DPW-RC, and CMU-Syn datasets. As illustrated in the visual examples (Figures~\ref{fig:3dpw_ori_picture}-\ref{fig:cmu_syn_1s3s_04}), our model yields more stable predictions in both  multi-agent dynamics and close-proximity interactions. In scenarios involving significant pose changes,  baseline methods tend to exhibit noticeable trajectory drift and pose deviations as the sequence progresses. Similarly, during close social interactions, we observe that baselines can produce misaligned limb positions or slight shifts in global translation compared to the ground truth. Our approach maintains a  closer overlap with the ground truth sequences (light dashed lines) over time. By effectively modeling the dynamic interactions, our method generates more accurate global trajectories and better preserves the  spatial positioning and pose details.


\begin{figure*}[!tp]
    \centering
    \includegraphics[
        width=\textwidth,
        height=0.75\textheight,
        keepaspectratio
    ]{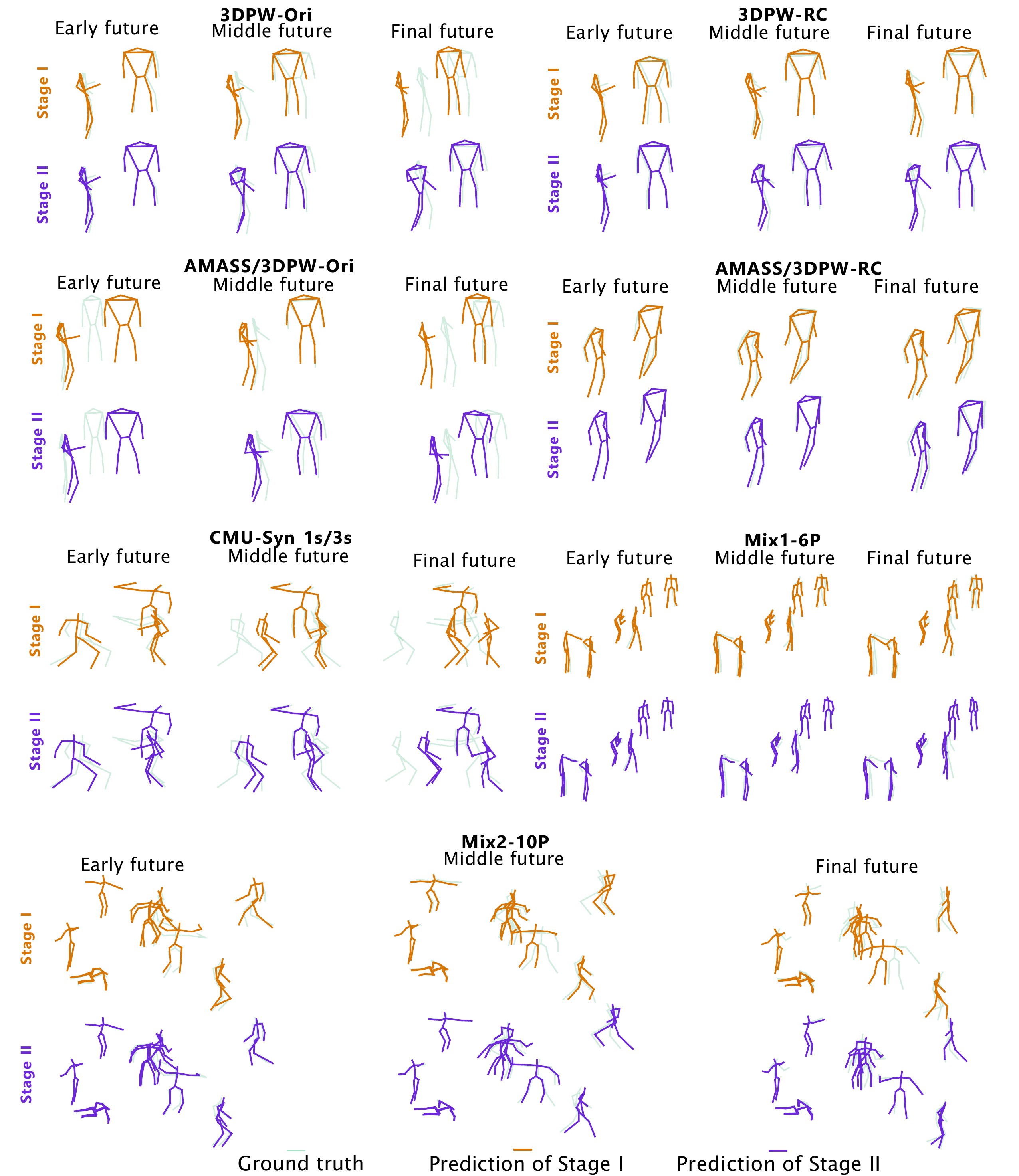}
    \caption{Qualitative comparison of our two-stage framework across diverse benchmarks, showing early, middle, and final future frames. Ground truth, Deterministic Coarse Prior (DCP, Stage I), and Residual Flow Matching (RFM, Stage II) are colored in light green, orange, and purple, respectively. The RFM refinement noticeably mitigates the trajectory drift and spatial misalignment present in the coarse prior.}
    \label{fig:two_stage_refinement}
\end{figure*}

Figure~\ref{fig:two_stage_refinement} further illustrates the proposed two-stage generative paradigm by comparing the predictions from the Deterministic Coarse Prior (DCP, Stage I in the figure) and the Residual Flow Matching refinement (RFM, Stage II). While the DCP module establishes a kinematic anchor and captures the primary motion trends (orange skeletons), it exhibits trajectory drift and spatial misalignment over extended prediction horizons (e.g., the ``Final future'' frames). By formulating the continuous-time flow over motion residuals, the RFM module mitigates these deviations (purple skeletons). Guided by the Dynamic Cross-Interaction (DCI) mechanism to incorporate social contexts, the refinement process helps align both global trajectories and local body articulations closer to the ground truth (faded light blue). These structural improvements can be observed across diverse settings, including original and camera-motion-compensated coordinates (3DPW-Ori and 3DPW-RC), extended forecasting horizons (CMU-Syn 1s/3s), and highly crowded environments (Mix1 (6-person) and Mix2 (10-person)).


\end{document}